\documentclass[graybox]{svmult}

\usepackage{graphics}
\usepackage{mathptmx}       
\usepackage{helvet}         
\usepackage{courier}        
\usepackage{type1cm}        
\usepackage{makeidx}         
\usepackage{graphicx}        
\usepackage{multicol}        
\usepackage[bottom]{footmisc}

\RequirePackage[textsize=scriptsize]{todonotes}

\makeindex             

\begin{document}

\title*{How did the discussion go: Discourse act classification in social media conversations}
\titlerunning{Discourse Act in Social Media}
\author{Subhabrata Dutta, Tanmoy Chakraborty and Dipankar Das}

\institute{Subhabrata Dutta \at Jadavpur University, Kolkata, India, \email{subha0009@gmail.com}
\and Tanmoy Chakraborty \at IIIT Delhi, India, \email{tanmoy@iiitd.ac.in}
\and Dipankar Das \at Jadavpur University, Kolkata, India, \email{dipankar.dipnil2005@gmail.com}}
%
%
\maketitle

\abstract*{}

\abstract{Within last two decades, social media has emerged as almost an alternate world where people communicate with each other and express opinions about almost anything. This makes platforms like Facebook, Reddit, Twitter, Myspace etc. a rich bank of heterogeneous data, primarily expressed via text but reflecting all textual and non-textual data that human interaction can produce. 
We propose a novel attention based hierarchical LSTM model to classify discourse act sequences in social media conversations, aimed at mining data from online discussion using textual meanings beyond sentence level. The very uniqueness of the task is the complete categorization of possible pragmatic roles in informal textual discussions, contrary to extraction of question-answers, stance detection or sarcasm identification which are very much role specific tasks. Early attempt was made on a Reddit discussion dataset. We train our model on the same data, and present test results on two different datasets, one from Reddit and one from Facebook. 
Our proposed model outperformed the previous one in terms of domain independence; without using platform-dependent structural features, our hierarchical LSTM with word relevance attention mechanism achieved F1-scores of 71\% and 66\% respectively to predict discourse roles of comments in Reddit and Facebook discussions. Efficiency of recurrent and convolutional architectures in order to learn discursive representation on the same task has been presented and analyzed, with different word and comment embedding schemes. Our attention mechanism enables us to inquire into relevance ordering of text segments according to their roles in discourse. We present a human annotator experiment to unveil important observations about modeling and data annotation. Equipped with our text-based discourse identification model, we inquire into how heterogeneous non-textual features like location, time, leaning of information etc. play their roles in charaterizing online discussions on Facebook.}

\section{Introduction}
\label{sec1}
While the predominant mode of people engaging in discussions on social media is via text, recent advents have pushed these communication to non-textual modes also. The most simple example of such communications are \textit{reactions} in Facebook. People tend to express their opinions towards a content or opinions of others via categorized reactions like `love', `angry' or a mere simple `like'. Platforms like Reddit provide `upvote' and `downvote' options to express categorized opinions. In an ongoing discussion, these options add newer structures of repartee discourse; users tend to express their opinion towards others using these options and  such non-textual mode of discourse run parallel to the textual one.
\par There are still other factors effecting how online discussions proceed. There are heated topics of discussion where an user represents his/her community sentiments, like religion, race, political affiliation, location etc. People bunk onto the sources of information like news reports, videos and images and these behaviors change with time. Even for a single topic, character of discussions varies temporally as new information floods in, from multiple modes.  Exploring relationships between these heterogeneous features may reveal valuable understanding of online discussions. But to continue, one needs to identify intentions or roles of different people in discussion, depending on text only. Heterogeneity can be explored over that primary identification of discourse.
\par The term \verb|Discourse| has been defined in numerous ways in linguist community. Broadly, discourse is how we meaningfully relate written or spoken natural language segments. In case of dialogues, we deal with compound discourses constituted by \verb|Narrative Discourse| and \verb|Repartee Discourse| \cite{larson1984meaning}. While narrative discourse focuses on the depiction of motion, repartee discourse engages to describe speech exchanges. This second part varies in nature with varying types and platforms of dialogue. For example, in spoken dialogues, utterances of a single speaker are much more intervened compared to email conversation, due to more interruption and real time transmission-reception when we talk face-to-face.
\par With the boom of social media, more and more people are expressing opinions, queries and arguments on topics innumerable, opening a completely new type of repartee discourse. This has opened scopes of understanding how people engage in discussions. In fact, this can be extended to almost any platform where people interact with each other in an informal manner like Facebook, Twitter, Reddit, CreateDebate etc. Participation in discussions are not homogeneous across these platforms; while Facebook or Twitter are more used for expression of opinions and argumentation, platforms like Reddit or different community forums have large usage for querying and answering. One method of understanding discussions has been to identify high-level discourse structures in such conversations. These structures tend to assign categories called \verb|Discourse Acts| to each textual utterance (comments, messages etc.) that pertain to their role in the conversation. 
\begin{table}[h]
\caption{Examples of Discourse Acts in discussion threads}
\label{tab:1}       
%
%
\begin{tabular}{|p{9cm}|p{2.5cm}|}
\hline
\textbf{Comment} & \textbf{Discourse Act Tag} \\
\hline
\textbf{U1}: \textit{I'm not form the US but am interested in US politics. So here is my question: Is Obamacare failing? If the democrats had won. Would they also be in a hurry to fix Obamacare because of flaws in the law?} & \textbf{QUESTION}  \\
\hline
\textbf{U2}(replying \textbf{U1}): \textit{The straight up answer to your question is "it's complicated". In my point, million have obtained insurance, some who were uninsurable before, primary care is available to more, maternal care, mental health, and more are available to closer to everyone in the US.} &     \textbf{ANSWER} \\
\hline
\textbf{U3}(replying to \textbf{U2}): \textit{But not as many people obtain insurance as the CBO predicted in 2009, insurance costs too much, and the Obamacare is still a massive government overreach.} &  \textbf{DISAGREEMENT}  \\
\hline
\textbf{U4}(replying to \textbf{U2}): \textit{That was a good answer!} &  \textbf{APPRECIATION}  \\
\hline
\end{tabular}
\end{table}  
\par Like discourse processing in plain documents, discourse parsing of dialogues is a two step process: identification of discourse constituent or elementary discourse units and then establishing discourse relation between them. In dialogues, each utterance is linked to the utterance it was replied to. In Table~\ref{tab:1} \textbf{U2} is linked to \textbf{U1}, \textbf{U3} is linked to \textbf{U2} and so on. In case of structured platforms like CreateDebate or Reddit, these links are already known. Each comment in these platforms is an explicit reply to some other. But in Facebook or various group chat platforms, this structure is not explicitly known. Dutta et al. \cite{dutta2017dialogue} proposed a Support Vector Machine based framework to decide which comment is put in reply to whom, in case of Facebook discussion threads. A much complete work on tagging discourse acts of discussion comments is the \verb|Coarse Discourse Dataset| \cite{zhang2017characterizing}. This is a Reddit dataset with over 9000 discussion threads comprised of over 100000 comments. Each comment is classified into one of nine different discourse act tags, namely, \textbf{Announcement}, \textbf{Question}, \textbf{Answer}, \textbf{Elaboration}, \textbf{Humor}, \textbf{Agreement}, \textbf{Disagreement}, \textbf{Appreciation}, and  \textbf{Negative reaction}, with undecidable roles as \textbf{Other}.
\par Given the constituency information, a single discussion thread becomes a tree with the first or opening comment being the root node. Depth-first traversal of this tree yields multiple linear sequences of comments; each posed as a reply to its previous one. We then hypothesize that, identification of discourse role of each comment depends on its ancestors along the chain and not only its parent comment; thus the problem becomes classification of sequence of comments to corresponding sequence of discourse acts. This is a familiar problem of mapping input sequences to same length output sequences, and a variety of traditional and neural learning models exist to solve this.
\par \verb|Reccurent Neural Networks| or RNNs revolutionized the modeling of sequential data with its emergence. Given an RNN with $x_t$ as input in current timestep, $h_{t-1}$ as hidden layer output from previous timestep, output for current timestep $o_t$ is computed as $f(x_t,h_{t-1})$ where the network learns function $f$. Theoretically, this enables an RNN to learn dependencies in sequences which can be spread to arbitrary distances. Practically this is not the case, as \textit{Vanishing Gradient Problem} \cite{hochreiter2001gradient} restricts RNNs to learn long term dependencies. This is where \verb|Long Short Term Memory| or LSTM models \cite{hochreiter1997long} come into play. In LSTMs, a separate memory cell is used to remember long term dependencies, which can be updated depending on current input; So at each timestep, LSTM takes current input $x_t$ and previous memory cell state $c_{t-1}$ as input and compute output $o_t$ and current cell state $c_t$. Governing equations of an LSTM are\begin{equation}
i_t = \sigma_i(x_tW_{xi}+h_{t-1}W_{hi}+b_i)
\end{equation}
\begin{equation}
f_t = \sigma_f(x_tW_{xf}+h_{t-1}W_{hf}+b_f)
\end{equation}
\begin{equation}
\label{celleq}
c_t = f_t\odot c_{t-1}+i_t\odot \sigma_c(x_tW_{xc}+h_{t-1}W_{hc}+b_c)
\end{equation}
\begin{equation}
o_t = \sigma_o(x_tW_{xo}+h_{t-1}W_{ho}+b_o)
\end{equation}
\begin{equation}
h_t = o_t\odot \sigma_h(c_t)
\end{equation}
where $x_t$ is input vector, $f_t$ is forget gate activation vector, $i_t$ is input gate activation vector, $o_t$ is output gate activation vector, $h_t$ is output vector of LSTM, $c_t$ is cell state and $W$ and $b$ are weight and bias matrices. Malhotra et al. \cite{malhotra2015long} showed that stacking up LSTM layers above each other enhances learning; deeper layers tend to model more complex representations from the previous layer output. This idea of stacking up multiple LSTM layers will be used in our experiments also. 
\par LSTMs have been proved to be very much efficient in NLP tasks where sequence classification or time series forecasting applies \cite{sundermeyer2012lstm,sutskever2014sequence,wang2016attention}. LSTMs, when allowed to focus particular segments of input data, perform even better. Intuitively, this directs the model to remember relations regarding focused segments with additional priority. This is the basic idea behind \verb|Attention Mechanisms|.
\par We pose our problem to identify discourse role of participants in online discussion with a broader problem of mapping heterogeneous real world phenomenons with the characteristics of discussions on social media. We organize the rest of the chapter in following manner:
\begin{itemize}
\item In Sec.~\ref{sec:related_work} we present a survey of works related to linking real word phenomenons with social media data, network analysis of social media interactions, discourse act tagging and an important subproblem of stance detection.
\item Sec.~\ref{sec:model_desc} presents the working and underlying rationale of our proposed multidimensional LSTM model with attention mechanism for discourse act tagging, along with one multi-layer perceptron model, two pure LSTM based models and one convolutional LSTM model.
\item With the model definitions completed, we move on to experiment methodologies used in detail in Sec.~\ref{sec:experiment}. 
\item Sec.~\ref{sec:observe} contains the evaluations of the model performance along with a comparative study.
\item We introspect into the shortcomings of our models and propose some possibilities of overcoming them in Sec.~\ref{sec:error}.
\item In Sec.~\ref{sec:utility} we discuss how our model can be used to characterize online discussions to predict temporal variation of argumentation, possible effects of external sources of information and emergence and reflection of community sentiments.  
\end{itemize}

\section{Related Work} \label{sec:related_work}
Linking social media to real world events has gained much focus from the start of this decade; mostly related to entity recognition and opinion mining. \cite{bollen2011twitter} analysed text contents of tweet streams to build a \textit{mood time series} and studied its relation with the stock market time series. They used two different mood tracking tools to tag each tweet, tested those tools to predict moods of user regarding two different social events and finally devised a Granger Casuality Analysis and a Self-Organizing Fuzzy Neural Network to find correlation between this mood time series and daily up-downs of Dow Jones Industrial Average (DJIA) from March 2008 to December 2008. They achieved an accuracy of $86\%$ to predict the direction of DJIA. Similar type of work was presented by Wang et al. \cite{wang2012system}; they proposed a real-time system for Twitter sentiment analysis of US presidential election 2012. O'Connor et al. \cite{o2010tweets} linked several surveys on political presidential poll and consumer confidence in US to contemporary Twitter text sentiment. Most of these works and related ones tend to analyse sentiments of opinions expressed in social media data. Tan et al. \cite{tan2011user} incorporated Twitter networking structure to classify user-level text sentiments.
\par As they become a platform to reflect thoughts and opinions, social media has its intrinsic role of networking, of connecting people and make some information flow. Subsequently, problems like community detection, information flow prediction, rumor detection etc. emerge. Chakraborty et al. \cite{chakraborty2017metrics} presented a survey of metrics to evaluate community detection systems. Without going into much detail, one  can refer to \cite{lotan2011mapping,suh2010want,wang2015bandwagon} as noteworthy works on analysis and prediction of what type of content propagates more over social media. 
\par Research developments discussed till now have mostly relied on the semantics of the text in focus. Analysing text in discourse level to do the same job is less explored yet promising approach. Trevithick and Clippinger \cite{trevithick2008method} proposed how speech acts of message contents can be used to characterize relationship between participants in social networking. Somasundaran et al. \cite{somasundaran2009supervised} showed how opinion polarity classification can be improved by considering discourse relations. 
\par The idea of discourse acts originated from spoken dialogues (also called \verb|speech| \verb|acts|). Bunt \cite{bunt2010methodology} proposed an ISO standardization for dialogue act annotation in spoken dialogues. Clark and Popescu-Belis \cite{clark2004multi} proposed MALTUS, a multi-layered discourse act tagging for spoken dialogues. In case of textual conversations they do not readily translate. Repartee discourse, as discussed in Section \ref{sec1} varies in qualitative nature when we go from speech to text, a major cause being the asynchronic nature of textual exchanges. Within the textual arena, formal communications like emails show distinctly different nature of exchange compared to informal conversations like chats or open discussion platforms. Assigning speech act like labels to emails was another approach \cite{cohen2004learning}; this labeling was based on an idea about intention of action expressed by the mail sender, so they used tags like \textit{request} or \textit{commitment} as discourse roles. Kalchbrenner and Blunsom \cite{kalchbrenner2013recurrent} proposed a neural model for classifying dialogue acts in transcribed telephone conversations. They used convolutional architecture to produce sentence representation from word embeddings, and then a recurrent network to map each convoluted utterance to corresponding discourse act. A similar type of architecture has been implemented in our work for comparison.
\par Most of the previous research in online discussion has focused on extraction of \textit{question-answer} pairs. Ding et al. \cite{ding2008using} developed a CRF based model to identify context and question-answer discourse in a dataset constructed from Tripadvisor forum. Some endeavored on understanding argumentative discourse in online platforms dedicated for debating \cite{hasan2014you,misra2017topic} and limited to a handful of topics. Bhatia et al. \cite{bhatia2016identifying} proposed a discourse act classification scheme on Ubuntu and Tripadvisor forum posts. Although they tend to classify not extract specific types of posts, the data they chose and discourse acts they specified was more of query-solution type. Arguello and Shaffer \cite{arguello2015predicting} handled a similar problem, predicting discourse acts for MOOC forum posts. Both of the previous works focused on similar type of data on limited domain: they tend to classify discourse roles of posts which are put to mostly ask for help, answer some queries or evaluate some previous post. These are not like new age social networks with almost no bar on topics of discussion, covering from political debate to simple query-answering related to restaurants. 

\par As mentioned earlier, a variety of approaches has been proposed regarding argumentation discourse, precisely, stance detection. Now stance detection can be viewed as a two-way problem. Both for monologues and dialogues, stance detection can be approached with fixed target set. That is, the subject about which to detect the stances are predefined. On the other hand, dynamic understanding of debate topic and stance detection around those topics is a much more complex task to solve. O’Connor et al. \cite{o2010tweets} presented a stance detection model from a large Tweet dataset covering numerous topics. Identification of political stances in social have gathered much focus recently. With a fixed target subject, Lai et al. \cite{lai2016friends} and Wang et al. \cite{wong2013media} provide much insight to this problem.
\par The attention based LSTM model proposed by Du et al. \cite{du2017stance} bears much similarity with our proposed model. They addressed the problem of target specific stance detection using neural models. To make an LSTM focus on parts of text which may relate to the target subject, they used word vectors augmented with vectors representing words of target topic.
\par All these works mentioned earlier focused on specific types of discourse, and relied on data at par with those discourse types. To our best knowledge, Zhang et al. \cite{zhang2017characterizing} proposed the first complete discourse categorization of textual discussions in a broad platform, dealing with numerous topics. They proposed a CRF based model to predict high level discourse act labels of comments, using textual content based features as well as structural features. With all the features, this model achieved a F1 score of \textbf{0.747}. But without structural features, F1 score dropped to \textbf{0.507}. The structural features they used were word counts in a comment, depth of comment in thread, length of sentences etc. These features are highly dependent on which discussion forum is being referred to. Their dataset was prepared from Reddit discussion threads, and therefore these structural features predominantly correspond to Reddit's own discussion types. For example, if one plans to test this model on Twitter discussion, structural features will not translate due to word limit of tweets. Same goes for Facebook as users idea about engaging in conversations in Facebook differs from Reddit, thereby changing the way people talk. Scott et al. \cite{scott2015pragmatics} and Misra et al. \cite{eisenlauer2014facebook} presented two independent studies on discourse of social media, focusing on Twitter and Facebook respectively. Both of them show a common finding, mediator platform with its functionality and constraints largely determines the pragmatics of mediated conversations. That is why we use the coarse discourse dataset as primary data to train and test our models, but exclude explicit representation of structural features and focus solely on content to make our model platform independent.

\section{Model Description}\label{sec:model_desc}
As there is no previous neural network model for the high level discourse labeling tasks in case of asynchronous textual conversations, to the best of our knowledge, we devised \textbf{five} different neural network based models along with our finally proposed multidimensional LSTM model with word relevance attention mechanism. We begin by describing these models.
\begin{figure}
\centering
\includegraphics[width=0.8\textwidth]{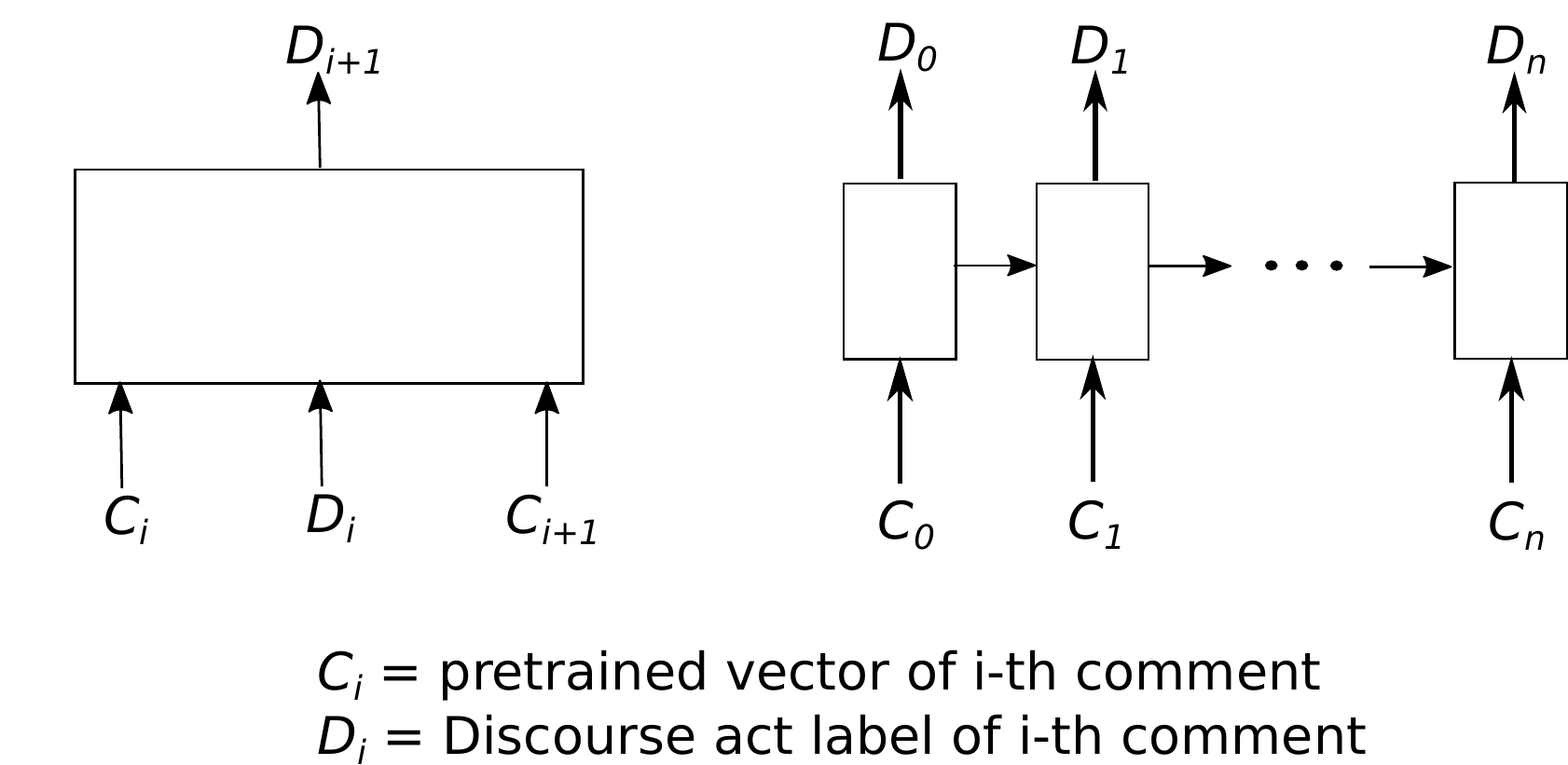}
\caption{Architectures of MLP(left) and LSTM(right) using pretrained comment vectors.}
\label{fig:MLP-LSTM}
\end{figure}
\subsection{Multi-Layer Perceptron Model}
In Section~\ref{sec1} we hypothesized that, prediction of discourse role of a comment in a chain is affected by all its ancestors in that chain. An MLP model contradicts this hypothesis, predicting discourse acts one by one. We present this model to justify our hypothesis. 
\par Let $C_i$ and $C_{i+1}$ be the pretrained vector representation of comments in a chain where $C_{i+1}$ is put in reply to $C_i$; $D_i$ is the discourse act represented in an one-hot vector of size $10$ (No. of classes). Input to our MLP is a vector $I$ resulting from concatenation of $C_i$, $D_i$ and $C_{i+1}$, and model predicts  $D_i+1$. Stepwise internal computations as follows:
\begin{equation}
\bf{H_1} = f_1(\bf{I}\bullet\bf{W_1} + \bf{B_1})
\end{equation}
\begin{equation}
\bf{H_2} = f_1(\bf{H_1}\bullet\bf{W_2} + \bf{B_2})
\end{equation}
\begin{equation}
\bf{D} = f_2(\bf{H_1}\bullet\bf{W_2} + \bf{B_2})
\label{eq:prob}
\end{equation}
where $f_1$ and $f_2$ are \textit{Sigmoid} and \textit{Softmax} nonlinearity respectively, $\bf{W}$ and $\bf{B}$ are weight and bias matrices of corresponding layer. Last output $\bf{D}$ gives a probability distribution over the $10$ classes to predict. the MLP tries to minimize the \textit{categorical cross entropy} of $\bf{D}$ in the course of learning.
\subsection{LSTM with Pretrained Comment Vectors}
Now to go along with our hypothesis, we design a simple LSTM model (Fig.~\ref{fig:MLP-LSTM}) with sequence of comment vectors $\mathbf{C} = [C_0, C_1,\dots,C_n]$ as input and sequence of predicted discourse acts $\mathbf{D} = [D_0, D_1,\dots,D_n]$ as output. Sequential outputs from the LSTM are connected to a densely connected layer with softmax activation to get the probability distribution of $D_i$'s, just like Eq.\ref{eq:prob}. Similar to the MLP, this model attempts to minimize \textit{categorical cross entropy} for each $D_i\in\mathbf{D}$.
\subsection{2-Dimensional LSTM}
\label{subsec:2DLSTM}
\begin{figure}[h]
\centering
\includegraphics[width=0.8\textwidth]{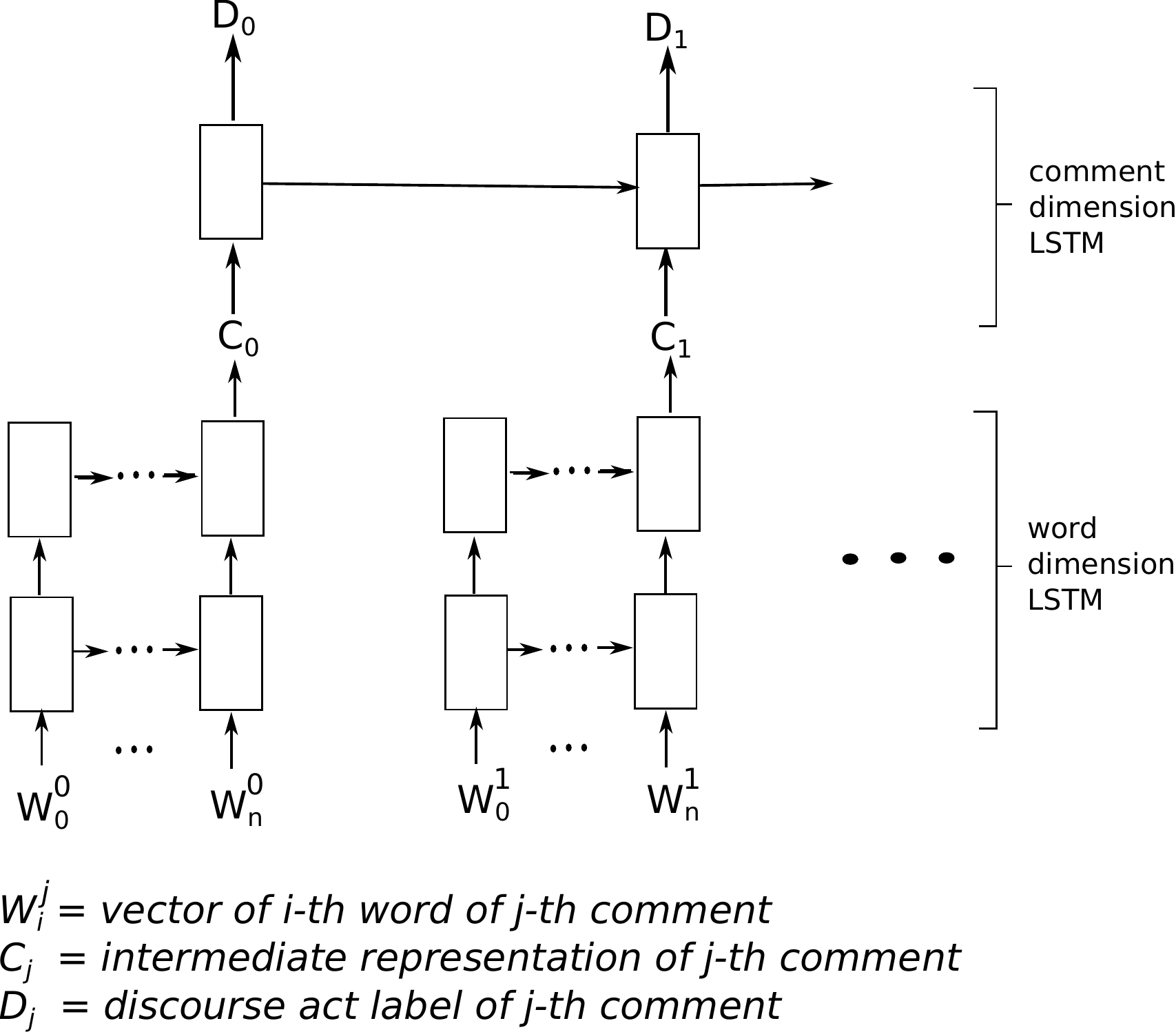}
\caption{Architecture two-dimensional LSTM}
\label{fig:2DLSTM}
\end{figure}

Previous two models take each comment as a pretrained vector. Our next model explore the task word by word and representations of comments are learned within discourse act prediction task. Rationale behind this was the assumption that, given the discourse act tagging target, intermediate representation of comments learned by sequential processing of words will contain more long distance dependencies between words. As we are dealing with pragmatic task, larger contexts of words are needed to be taken into account, and theoretically LSTMs can capture such dependencies better. As depicted in Fig.~\ref{fig:2DLSTM}, each comment $C_j$ is represented
as a sequence of words $\{W_{i}^{j}\}$. The model can be segmented into two parts: \textit{word-dimension stacked LSTM} and \textit{comment-dimension LSTM}. If we denote each LSTM layer as a black-box function $\mathcal{L}$, then the stepwise computations are as follows:
\begin{equation}
\mathbf{W'} = \mathcal{L}(\mathbf{W})
\label{l1}
\end{equation}
\begin{equation}
\mathbf{C} = \mathcal{L'}(\mathbf{W'})
\label{l2}
\end{equation}
\begin{equation}
\mathbf{D} = \mathcal{L}(\mathbf{C})
\label{l3}
\end{equation}
where $\mathcal{L}$ returns sequential output and $\mathcal{L'}$ returns single output. 
\par We used one-hot vector representation for the input words; an embedding layer was used to produce distributed word vectors. Weights of this embedding layer was set to pretrained word vectors, so that, during training our model fine tunes the word vectors.
\subsection{2-Dimensional LSTM with Word Relevance Attention}
Compared to the MLP model, those using LSTMs approach the problem of discourse act prediction task as a sequence-to-sequence modeling. In two-dimensional LSTM model, we hataken into account how the comments are being constructed from sequence of words to capture pragmatic relation between words. Still, in two-dimensional LSTM model we just discussed, intermediate representation of each comment depends only on the words constituting itself. This bars the model to pick relevant words in relation to the previous comments, topics of discussion etc., and simply same words in two different comments with two different discourse context gets same relevance. We propose an attention mechanism with the two-dimensional LSTM model to let it learn to assign more relevance on particular words depending on context.
\subsubsection{Word Relevance Attention}
Apparently question-answer discourse is easy to identify. Questions possess distinct parts-of-speech ordering (mostly starting with \textit{verbs}), and answers are always paired with questions. Discourse roles like agreement, disagreement, humor etc. are rather complex to distinguish. These discourse relations can often be identified with stances and content to justify stances. These stances can be based on the topic of discussion or the parent comment. That is, the task becomes to identify stance of two comments against a set of common topic words. We exploit a special structural leverage of online public forum discussion to identify topic of discussion. In case of human-human interaction via speech or via personal messages, usually a person is able to recall only a last few utterances. But in public forums, all the previous comments are open to read. Out off all those messages, the first (thread starter) comment is the one defining the topic of discussion. Every user tries to post their comment in relevance to those topics. 
\par The attention mechanism we propose is hypothesized to exploit these phenomenons. We select specific words(\textit{nouns}, \textit{verbs}, \textit{adjectives} and \textit{adverbs}) from the first comment of the thread and the parent comment (one that is replied to) as \textit{discourse target}. We take weighted mean of these words to produce two vectors, corresponding to the first comment and the previous comment(in our experiment we used \textit{tf-idf} weights to focus on important words from these two comments). These two vectors are then augmented with each word of the current comment. If $\mathbf{W_f} = [w^{f}_0,\dots,w^{f}_{M-1}]$ and $\mathbf{W_p} = [w^{p}_0,\dots,w^{p}_{N-1}]$ be the sequence of words extracted from the first and the previous comments respectively, then 
\begin{equation}
W_{f}'= \frac{\sum\limits_{i=0}^{M-1}{w^{f}_{i}t^{f}_i}}{M}
\end{equation}
\begin{equation}
W_{p}'= \frac{\sum\limits_{i=0}^{N-1}{w^{p}_{i}t^{p}_i}}{N}
\end{equation}
where $t^{f}_i$ and $t^{p}_i$ represents tf-idf value of the $i-th$ word in $\mathbf{W_f}$ and $\mathbf{W_p}$ respectively. We do not use plain tf-idf values calculated with comments being represented as documents. Instead, as proposed in \cite{dutta2017dialogue}, we use hierarchical frequencies. As the first comment represents topic of discussion, and this is a characteristic of the thread, we take threads as documents and calculate \textit{inverse thread frequency} of the words. With the previous comment words, we rather focus on the discussion context, which changes comment by comment. To extract relevant context words, for the previous comment, we compute \textit{inverse comment frequencies} over the whole dataset. So $t^{f}_i$ is term frequency multiplied by inverse thread frequency, whereas $t^{p}_i$ is term frequency multiplied by inverse comment frequency.
\par $W_{f}'$ and $W_{p}'$ now can be visualized as vectors representing relevant contents of first and previous comment respectively. They are concatenated with each word-vector of the current comment to generate target augmented vectors $\mathbf{T} = [T_0,\dots,T_C]$ where $C$ is the number of words in current comment. 
\begin{figure}
\centering
\includegraphics[width=0.6\textwidth]{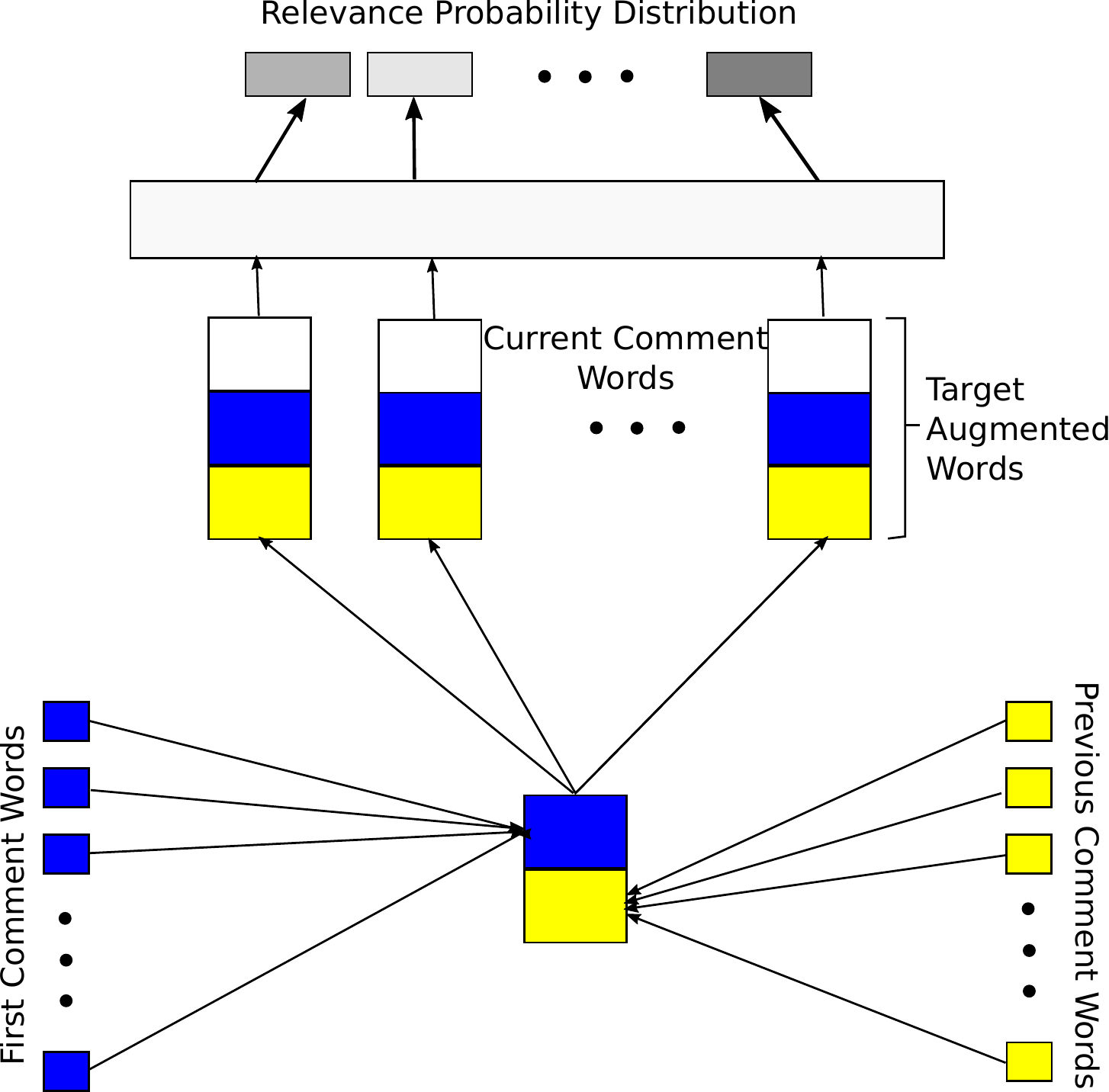}
\caption{Word relevance attention architecture.}
\label{fig:attention}
\end{figure}
\par Now we are going to make our model learn relevance ordering from $\mathbf{T}$. Given dimensionality of the word vectors used $V$, we initialize a trainable weight matrix $\mathbf{K}$ of order $C\times 3V$, and produce a vector ${S} = [s_0,\dots,s_C]$ such that,
\begin{equation}
s_i = \sum\limits_{j=0}^{3V}{\mathbf{K}[i][j]*\mathbf{T}[i][j]}\qquad \forall i\in [0,C]
\end{equation}
This operation serves two purposes. Suppose, $\mathbf{K}[i]$ contains all ones and all the word vector entries of $\mathbf{T}[i]$ are positive. Then larger values of $s_i$ will signify greater similarity between target word vector and $i-th$ word-vector of current comment. Practically word vector entries will be both positive and negative, and the attention may learn corresponding weight values in $\mathbf{K}$ to maximize $s_i$ for similar words. On the other hand, $mathbf{K}$ may learn to pose proper weights over related adjectives or negation words and even discourse connective prepositions to reflect their relevance when generating comment vectors. Applying \textit{softmax} over $S$, we get a probability distribution $P = [p_0,\dots,p_C]$ over the words of current comment.
\par Now we get back to two-dimensional LSTM model discussed in Section \ref{subsec:2DLSTM}. The relevance probability $P$ is applied to the outputs of the first LSTM of the word dimension part,
\begin{equation}
\mathbf{W_{rel}}[i] = p_{i}\mathbf{W'}[i] \qquad \forall i\in [0,C]
\end{equation}
where $\mathbf{W'}$ corresponds from Eq.~\ref{l1} and $\mathbf{W_{rel}}$ is the focused word sequence representation which will be fed to the next LSTM layer to generate comment vector.
\subsection{Convolutional Generation of Intermediate Comment Vectors}
In our proposed two-dimensional LSTM model, intermediate representation of each comment was computed using a stacked LSTM model. To compare the performance of this model, we also devised a convolutional model which generates a single vector from the \textit{word images} (sequence of words now constituting a 2D matrix of size $C\times V$) by convoluting and average-pooling. This model actually resembles to an \textit{N}-gram model, as we use three parallel 1D convolution with filter sizes two, three and four and concatenate the three pooling results to a single vector representing the comment. So the model actually learns to extract bi-gram, trigram and 4-gram features from the comments to generate the vector. Convoluted and concatenated vector is then fed to an LSTM to predict the discourse act sequence, identical to the comment-dimension LSTM of Section \ref{subsec:2DLSTM}. This model is experimented both with and without the attention mechanism proposed.
\begin{figure}
\centering
\includegraphics[width=0.6\textwidth]{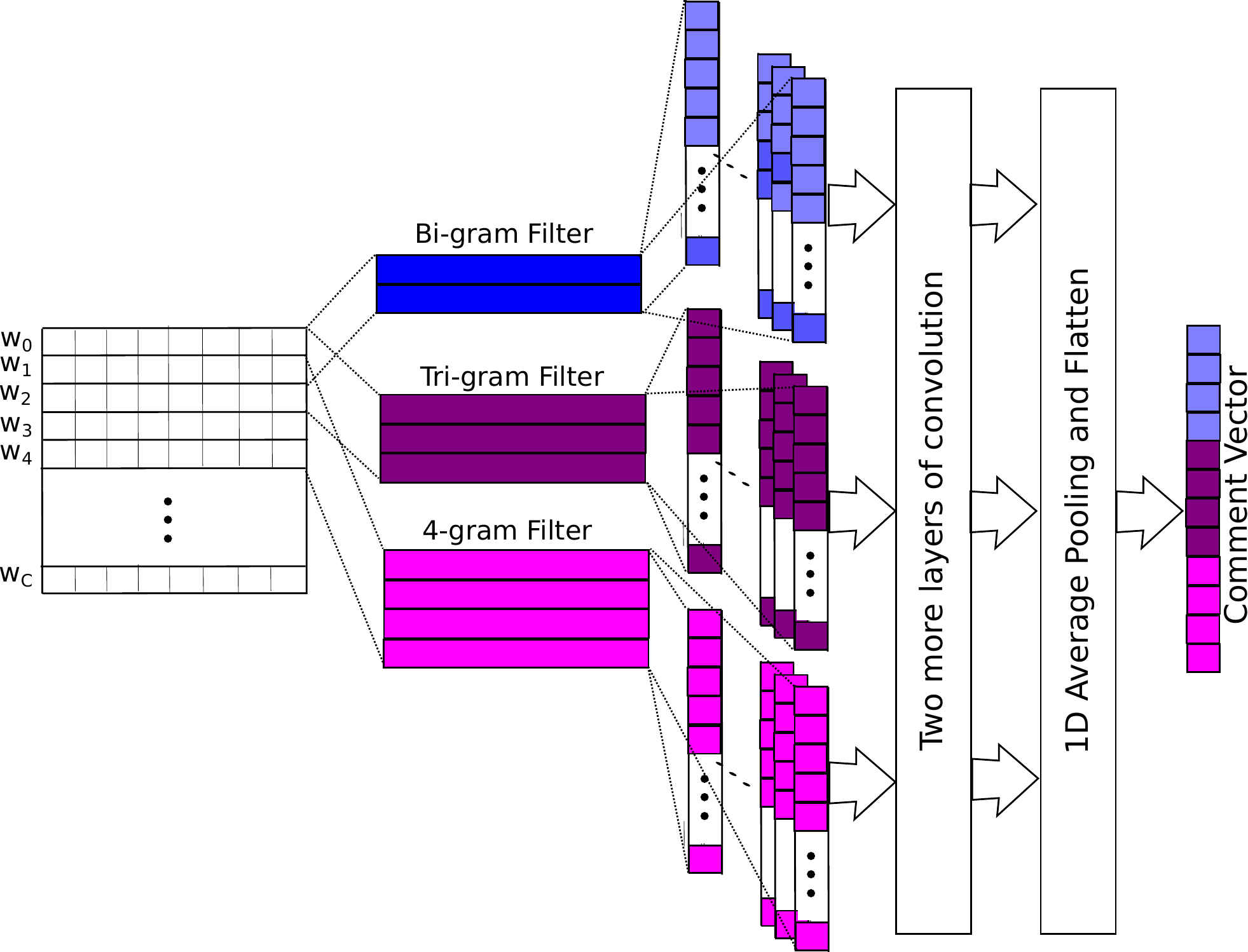}
\caption{ Architecture of convolutional generation of comment vectors(output goes to LSTM).}
\label{fig:cnn}
\end{figure}

\section{Experiments}\label{sec:experiment}
\subsection{Dataset}
We used the the \textit{Course Discourse Dataset}\footnote{https://github.com/dmorr-google/
coarse-discourse} of Reddit discussions as our primary dataset to train and test our models. An in-depth analysis of the data can be found in \cite{zhang2017characterizing}.
\begin{table}
\begin{center}
\begin{tabular}{l|c|c}
\hline Discourse Act & Total No. of Comments & \% in dataset  \\ \hline   
Question & 17681 & 17.6 \\ 
Answer & 41658 & 41.5 \\
Announcement & 2024 & 2.0 \\
Elaboration & 18927 & 18.8 \\
Agreement & 5072 & 5.1 \\
Disagreement & 3436 &  3.4 \\
Humor & 2409 & 2.4 \\
Appreciation & 8807 & 8.8 \\
Negative Reaction & 1899 & 1.9 \\
\hline
\end{tabular}
\end{center}
\caption{ Presence of different discourse act classes in Coarse Discourse Dataset }
\label{tab:discourse_act_class}
\end{table}
\par Another dataset of comparably smaller size was also used, solely to test the best three models from test results on the coarse discourse dataset. This dataset is an extended version of the Facebook discussion dataset prepared by Dutta et al. \cite{dutta2017dialogue}. We manually annotated the data with discourse act labels as per the rating guidelines with the coarse discourse data. This dataset contains 20 threads of discussions under posts from Facebook pages of various newsgroups like BBC, The New York Times, The Times of India, The Guardian etc., with a total 1177 comments altogether. These discussions were mostly related to heated contemporary social and political issues like \textit{US Presidential Election 2016}, \textit{Terrorist Attacks}, \textit{Brexit} etc..
\subsection{Pretrained Embeddings}
As discussed in the model descriptions, we used either pre-trained word or comment embeddings. For word embeddings, we used Word2Vec \cite{mikolov2013distributed} on the comments from both the datasets taken together. We generated six different word embedding sets with vector size $50$, $100$, $150$, $200$, $300$ and $400$ and ran pilot experiments with the two-dimensional LSTM and convolutional LSTM models using each embeddings on a small fraction of the dataset. We observed that increasing vector size beyond $150$ did not bring any further betterment of performance and only burdened the hardware performance. So we stick to the $150$-length word embedding throughout the experiments. 
\par Same idea was used for pretrained comment embeddings. Doc2Vec \cite{le2014distributed} was used to produce ten different paragraph embeddings ranging from size $500$ to $1500$,  and finally the one with size $700$ was taken.
\subsection{Training models}
We used 5-fold stratified cross validation to train and test all the models. As the length of comment sequences varies from 2 to 11, and further number of words present in each comment varies substantially, we had to pad each comment with zeros. Longer chains are actually rarer, and thus to avoid too much padded data, we take equal length sequences at a time. That is, we train a model on sequences of length $i$, save the learned weights and load them to a new model for sequences of length $i+1$ and so on. We optimized hyper-parameters of all the models using scikit-learn's\footnote{http://scikit-learn.org} Grid-search. 
\par A CRF model using all the features (content + structure) used by Zhang et al. \cite{zhang2017characterizing} was also trained. We test this model on Facebook dataset to present a full comparison of our models along with the state of the art.
\subsection{Human Annotator Testing}
From the \textit{inter-annotator reliability} measures presented by Zhang et al. \cite{zhang2017characterizing}, one can see that distinguishing discourse act labels other than question, answer and announcement is a very much subjective process. So we devised an evaluation of model performance by human subjects from linguistic and non-linguistic backgrounds. A small fraction of the testing data from Reddit dataset was used for this purpose(20 threads containing a total 657 sequences), with the same rating guidelines. Two different set of experiments were done using each group of annotators:
\begin{enumerate}
\item \textbf{Zero knowledge testing of model}. Annotators are presented with predictions made by a model. They evaluate the predictions without prior knowledge of the labeling in corpus.
\item \textbf{Zero knowledge testing followed by allowed correction}. Annotators do zero knowledge testing; then we present them the corresponding labeling in corpus. They are allowed to make changes in their previous decisions if they think so.
\item \textbf{Self-Prediction}. Annotators were asked to predict discourse tags, and we evaluated their performance compared to the corpus labels.
\end{enumerate}
\section{Observations}\label{sec:observe}
\subsection{Model Performances}
We start with presenting results of testing all the models proposed on coarse discourse dataset. As shown in Table~\ref{tab:all_result}, our first hypothesis of modeling the problem as a seq2seq prediction task clearly holds true. All four models processing sequence of comments as input outperforms the MLP model by big margin. We do not use this model for further experiments.
\begin{table}
\begin{center}
\begin{tabular}{l|c|c|c}
\hline Models & Precision & Recall & F1 score \\ \hline
MLP  & 0.51 & 0.43 & 0.46 \\ 
LSTM with paragraph-vectors & 0.58 & 0.57 & 0.57 \\ 
2D LSTM & 0.60 & 0.61 & 0.60\\ 
CNN-LSTM & 0.62 & 0.60 & 0.61\\ 
CNN-LSTM with attention & 0.65 & 0.63 & 0.64\\ 
\textbf{2D LSTM with attention} & \bf0.72 & \bf0.71 & \bf0.71\\ 

\hline
\end{tabular}
\end{center}
\caption{ Overall performance of all models tested on Reddit dataset. }
\label{tab:all_result}
\end{table}
\par
\begin{figure}[h]
\centering
\includegraphics[width=0.6\textwidth,height=0.6\textwidth]{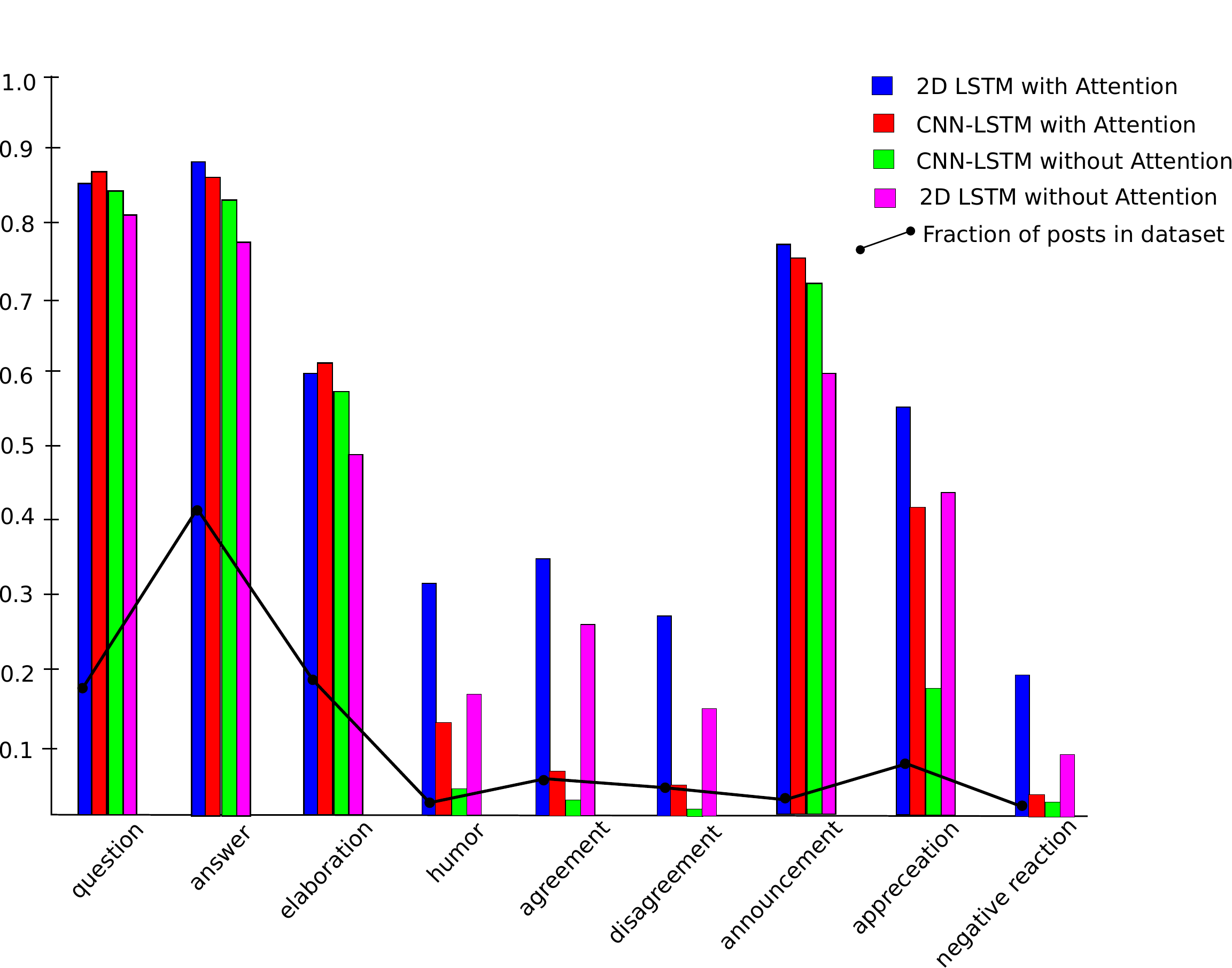}
\caption{Class-wise F1-scores of the four best performing models.}
\label{fig:allClass_allModel}
\end{figure}

 In an over all performance, convolutional generation of comment representation had an edge over recurrent generation when not equipped with the word relevance attention. To dig a bit further, we need to look into the class-wise performance of the four models presented in Fig.~\ref{fig:allClass_allModel}. The CNN-LSTM model actually performed much better compared to the 2D LSTM in classes like question, answer, announcement and elaboration. But classes like humor, agreement, disagreement, appreciation and negative reaction shows the other way round. Though this phenomenon demands an in-depth analysis, we present a rather intuitive explanation here.
\par Our CNN-LSTM model takes bi-grams, tri-grams and four-grams from the comment consecutively and constructs parallel representations based on them. This captures the local organization of text very well. Discourse roles of the first type are more easily predictable with such local organizations. But they fail to capture long distance relationships between words in a text, which is more important to identify argumentative discourses of the second type. LSTMs have an edge over CNN's when the problem is to identify long dependencies. 
\begin{figure}[h]
\centering
\includegraphics[width=0.6\textwidth,height=0.6\textwidth]{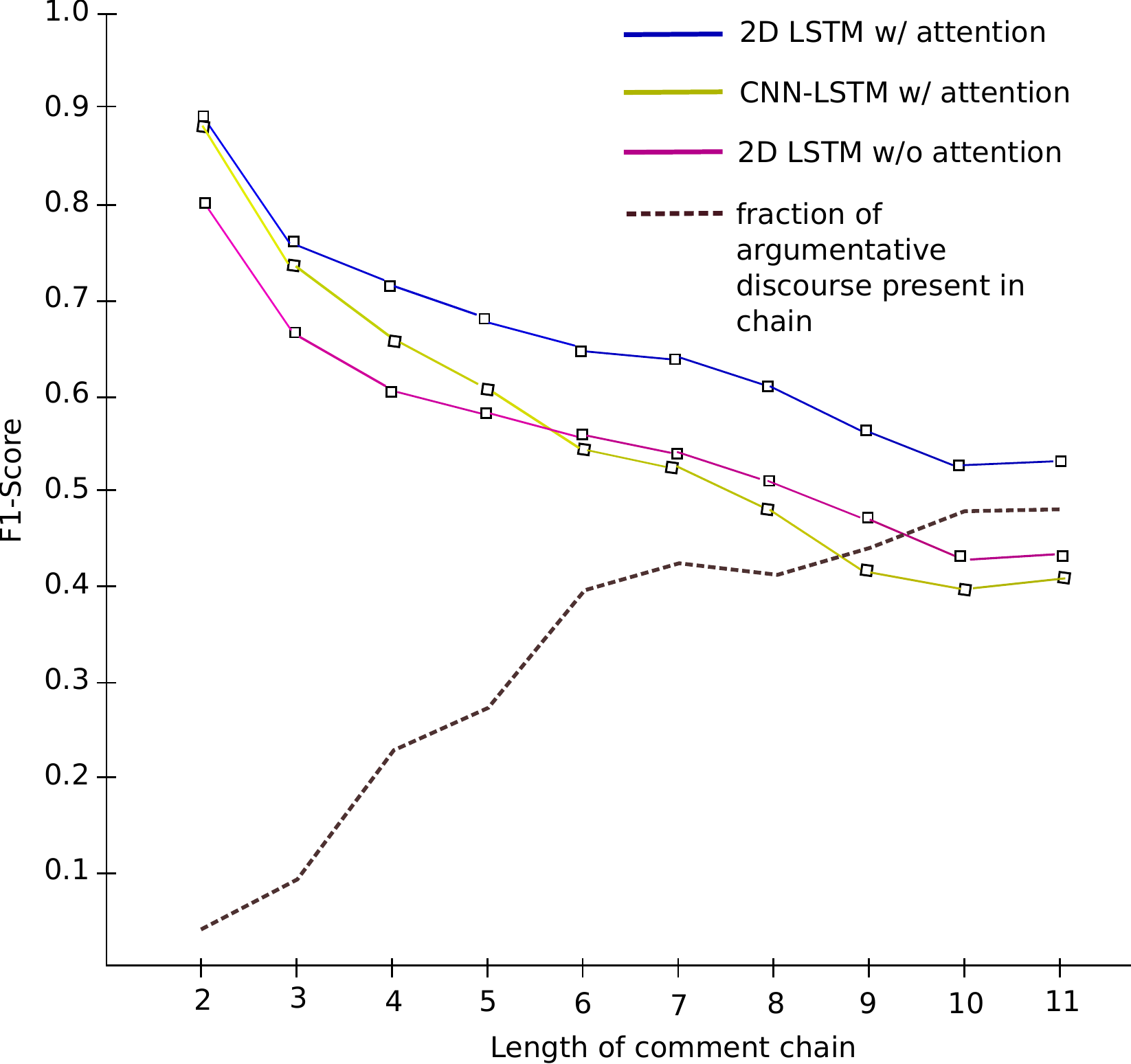}
\caption{Model performances and fraction of argumentative discourse roles present in sequence vs. length of comment sequence.}
\label{fig:chainwise}
\end{figure}
\par But simple 2D LSTMs does not learn to know which long term dependencies are actually to put more focus on. Word-dimension part of the model tries to capture internal discourse organization of a comment without taking into account what the high level discourse is going on. Theoretically, making these LSTMs stateful would have solved this problem, but in reality mere statefulness can not capture this much information content. The attention mechanism we proposed tried to solve this problem, at least partially. We can see the sheer rise of performance in both the CNN-LSTM and 2D-LSTM model when equipped with the word relevance probability computed using previous and first comment words. But here again, 2D LSTMs exploited the attention much more rigorously compared to the convolutional one.

\par From the nine discourse act roles, we identified \textit{agreement}, \textit{disagreement}, \textit{humor}, \textit{appreciation} and \textit{negative-reaction} as part of argumentative discourse. In Fig.~\ref{fig:chainwise}, presence of argumentative discourse can be seen more manifesting in longer chains. As the length of chain increases, two challenges occur simultaneously; the comment dimension LSTM have to remember the learnings from more previous comments; on the other hand, with the increase in argumentative discourse, stance detection problem with dynamic targets come more into play. We can check the performances of our two dimensional LSTM model, with and without word relevance attention, and CNN-LSTM with word relevance attention, given the variation of the two challenges.
\begin{table}
\begin{center}
\begin{tabular}{l|c|c|c}
\hline Models & Precision & Recall & F1 score \\ \hline   
CNN-LSTM with attention & 0.61 & 0.57 & 0.58\\ 
CRF with all features & 0.62 & 0.63 & \bf0.62\\
2D LSTM with attention & 0.65 & 0.68 & \bf0.66\\ 
 
\hline
\end{tabular}
\end{center}
\caption{ Comparison with state-of-the-art model on Facebook data. }
\label{tab:test_fb}
\end{table}
\par Now we move on to compare our best two models (CNN-LSTM and 2D LSTM, both with attention) with the state of the art CRF based model by Zhang et al. \cite{zhang2017characterizing}. On the Reddit dataset, CRF model with only content features (with F1 score $0.50$) was clearly outperformed by our models by big margin. However, the one with all the features still got better results compared to us, with F1 score $0.74$. But on the Facebook dataset, as presented in Table~\ref{tab:test_fb}, our 2D LSTM with word relevance attention clearly performed better compared to the all feature CRF. Though this test is not exhaustive with data from other discussion platforms, we can still conclude that our model achieved better domain independence.
\subsection{Human Annotator Evaluation}
Low inter-annotator reliability presented for the Reddit dataset reflects substantially in the test results from human evaluation. As we can see, when presented with the annotations done by our 2D LSTM with attention model, human annotators evaluated those to be better performing compared to the actual results evaluated by corpus annotations. When presented with the corpus annotations, these annotators changed their decision to decrease th scores. As we can see, this change is much higher for annotators from non-linguistic backgrounds.  
\begin{table}
\begin{center}
\begin{tabular}{l|l|c}
\hline Annotator Background  & Evaluation method & F1 score \\ \hline   
Linguistics & Zero knowledge & 0.74\\ 
Linguistics & Zero knowledge with correction & 0.72\\
Linguistics & Self-prediction & 0.81\\ \hline
Non-Linguistics & Zero knowledge & 0.77\\
Non-Linguistics & Zero knowledge with correction & 0.71\\
Non-Linguistics & Self-prediction & 0.73\\  \hline
Actual Performance & & 0.70\\
\hline
\end{tabular}
\end{center}
\caption{ Results of human annotator testing. }
\label{tab:human_results}
\end{table}
\par When annotators were asked to predict the labels themselves, their performance also varied by a big margin. Average scores for linguistic and non-linguistics background annotators have been presented. Worst case to best case results varied from $0.68$(by a non-linguistic annotator) to $0.82$(by a linguistic annotator). We also observed them to tag same comment in different sequences with different discourse roles. 
\subsection{Learning Word Relevance}
\begin{figure}
\centering
\includegraphics[width=0.8\textwidth]{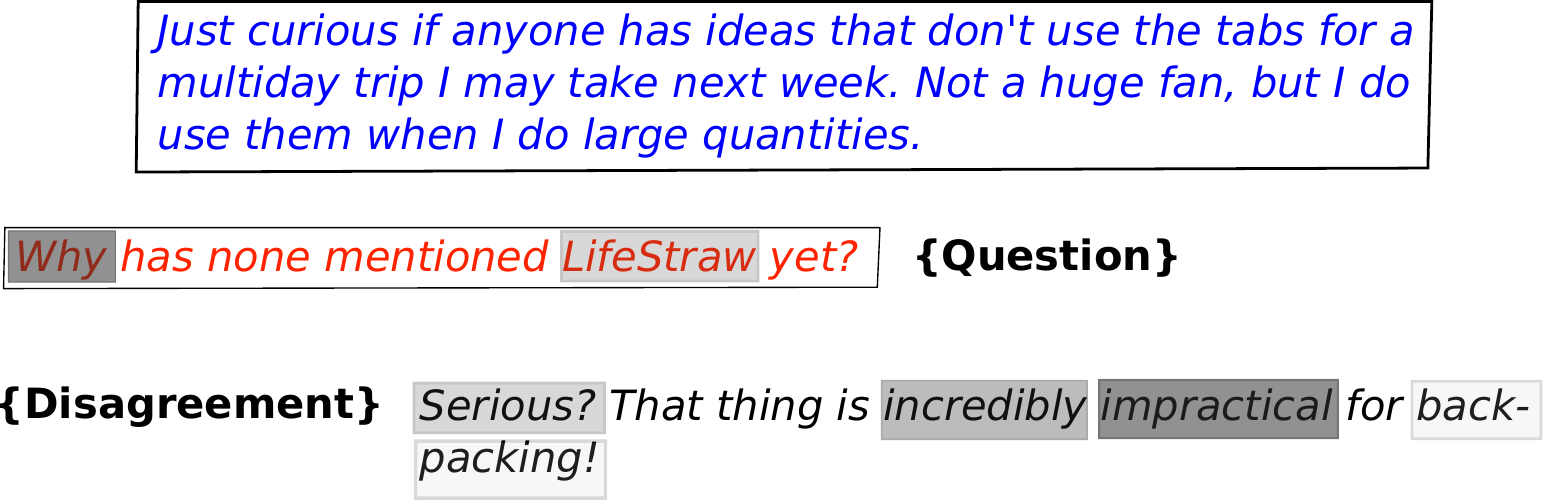}
\caption{Example of word relevance learned; blue text denotes first comment; darker labels on words signify higher relevance assigned.}
\label{fig:word_rel}
\end{figure}

Fig.~\ref{fig:word_rel} shows an example of how the attention mechanism weighted different words in the comment depending on the first and the previous comment. The comment shown in example acquired a disagreement-type discourse role. Darker gray labels signify higher relevance values assigned to the word by the attention mechanism. From the first comment we can see that the discussion evolves around some backpacking tools. The parent comment mentioned a named entity, possibly a particular brand name. The attention framework picked up the adjectives and adverbs in the current comment to as more relevant words. Also a discourse connective at the beginning was identified to mark the discourse role. For the parent comment also, words identifying the query nature of the comment has been put on more weighting.
\par From these relevance ordering, two related observations can be made. Firstly, with more fine tuning and a lot more of training data, this framework can lead us to an unsupervised marking of text for discourse connectives, importantly, high level discourse connectives. Unlike discourse connectives present in a single document, dialogue cues marking high level discourse might be expressed by word structures like some tree or graph. They do bear non-consecutive long distance interdependencies for sure. Secondly, for argumentative/stance-taking discourse, word relevance resembles sentiment lexicons. 
\section{Error Analysis and Possible Solutions}\label{sec:error}
Results of human annotator testing clearly indicates that labeling of discourse roles is very much a subjective process. In fact, a comment in a thread can have different discourse roles when taken in different chain of linked comments. Comments which manifest multiple discourse roles (e.g., disagreement and negative reaction simultaneously) should be allowed to be labeled with multiple tags with possibly a probability distribution over them. Or we can devise hierarchical labeling, so that each comment has a broad level discourse act tag, and more sub-acts under that.
\par We devised our attention mechanism to make the word-dimension LSTM peep into words from first and parent comment. But our very assumption that discourse act tagging is a seq2seq problem is not actually being exploited here. The flow of narrative discourse within comments in the chain other than the first and the parent one must be taken into account when the sequence of words is being modeled into a comment vector.
\par With the attention mechanism, our model was able to distinguish between query-type discourse and argumentative discourse. But within the these broad discourse types, every discourse classes bear some more complex semantic and pragmatic features, and a further focus is necessary for better prediction of similar discourse roles. For example, our model misclassified only 7\% of comments of query-type discourse as an argumentative one. But a huge 22\% of negative reaction-type comments have been misclassified as disagreement. Same goes for other similar classes. If we take disagreement vs negative reaction confusion as an example, focusing on textual features regarding justification or subjectivity would help us more. Humors are actually the most complex discourse to predict. In an argument, without the knowledge of the stance of a person beforehand, humors (to be precise, sarcasms) are almost similar to agreements or appreciations. Thus a speaker profiling would also be needed for the task. Coarse discourse dataset have an average 1.66 posts from each author. This is pretty high for a single dataset but not enough for a complex task like speaker profiling, specially using neural models. Possibly a joint source model can help here, where a separate author data will be collected and used in parallel.
\par Using pretrained word vectors incorporate semantic relations of words in a comment. For example, this makes the model to identify \textit{Democrats} being related to \textit{politics} or \textit{republicans} or \textit{Hilary}. But this can never capture the antagonistic relation between \textit{Obamacare} and \textit{Trump}. Thus, given two short texts with such words and not much explanation (discourse roles like negative reaction), a model would need either much larger training set or some way to incorporate world-knowledge. With our model, many of such instances have been misclassified.
\par Last but not the least, though we took a linear sequence of comments as input to classify, almost all discussion platforms originally follows a tree structure(the one from which we picked up each individual sequence). Now when a person starts typing a reply to a certain comment, he or she can see all the other comments present there; possibly many of those comments are reply to the comment he or she is writing a reply to. And these comments do have weak effects on discourse roles; that is, not only the linear sequence, the whole thread tree needs to processed at a time for better understanding of the discourse.
\section{Characterizing Discussions with Discourse Roles}\label{sec:utility}
Up until this point, we discussed our proposed model to classify discourse acts of comments from text. With our nine different discourse roles, we can, at least apparently, characterize discussions and explore temporal patterns, community sentiments etc. We separately collected 12 threads of discussion regarding \textit{Jallikattu} comprising 894 comments from 59 users. These 12 threads are the discussions on Facebook pages of different news groups posting reports and videos about Jallikattu. We used our hierarchical LSTM with attention to tag each comment with discourse roles and explore the following hypothesis: 
\begin{enumerate}
\item A set of threads with more question-answer or appreciation tags can be hypothesized to be one where participants are acquiring and sharing information. On the other hand, threads with disagreement, negative reaction or humor tags are more likely to be argumentative or quarrelsome one. Temporal changes in nature of discourse reflects changing engagement of people.
\item Disagreement, negative reaction or humor relation between two comments group the corresponding users to antagonistic communities; whereas agreement and appreciation reflect a belonging to similar community. 
\end{enumerate}
\begin{figure}
\centering
\includegraphics[width=0.8\textwidth]{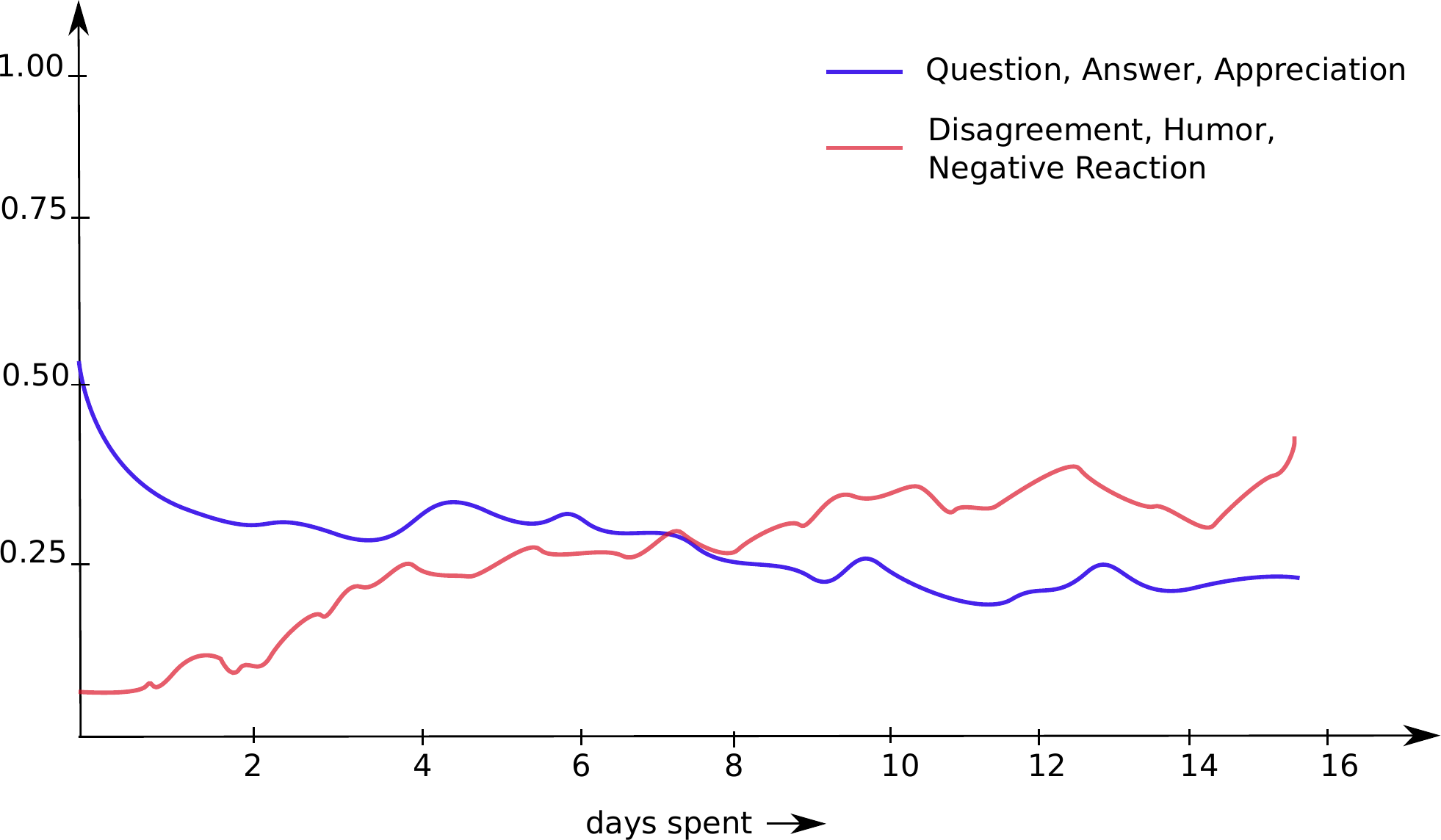}
\caption{Temporal plot showing how fraction of discourse patterns change with time spent over Jallikattu debate}
\label{fig:discourse_par_time}
\end{figure}
\par \textit{Jallikattu} is an ancient ritual in Tamilnadu (southern-most state of India), much like bull fighting. In 2016, a massive unrest followed the government decision to ban the festival. The decision was taken on 14th January. Fig.~\ref{fig:discourse_par_time} shows the change in discourse type in discussions over Facebook as time goes. One can clearly observe that, at the beginning the discourse showed an inquisitive nature, with mostly question-answer-appreciation type discourse acts prevailing. One explanation can be, within this time, people engaging in discussions were not much aware about what is happening and the focus was to know things. As time passes, mode of engagement changes more towards argumentation, with people taking firm position.
\par Our attempt to explore our second hypothesis revealed some actual community patterns. As stated, we tried to group people into two partitions depending on the discourse acts of their comments and to whom these were posted. For this purpose, we manually collected geographical location of the users participating in those discussions. Our observation was quite aligned with the real world situation: users almost got partitioned into two antagonistic groups, one mostly comprised of users from Tamilnadu, and the other mostly from northern India. 
\par A peculiar pattern about nature of argumentation revealed, depending on the original post under which the discussion is going on. When the original post was video or images of unrest, we got a sharp rise in two specific discourse acts, negative reaction and appreciation. In case of original post being extensive news report or socio-cultural analysis, elaboration, agreement and disagreement prevailed. The former can be specified as a case of subjective argumentation, with less objective reasoning, while the later may contain factual argumentation. 
\section{Conclusion}
Understanding complex pragmatics is a new and intriguing problem posed in front of computational linguists. With dialogues, this becomes far more challenging. In those work, we present a neural model which, naively speaking, employs layered memory to understand how individual words tend to constitute a meaningful comment, and then how multiple comments constitute a meaningful discussion. With the word relevance attention mechanism, our model tend to learn which words should be given more importance while deciding the discourse role of a comment. The proposed multidimensional LSTM with word attention not only outperformed the previous work on complete discourse act tagging, but also yielded better results in subproblems like question-answer extraction or stance detection compared to many previous works.
\par We can extend this idea of world relevance to larger contexts than only the previous and first comments. This may even equip us with an introspection into propagation of topic, sentiment and other linguistic features through conversations. We used the only available dataset with a sizable amount of training samples, and the imbalance of per-class samples might have restricted our model to achieve best performance. But with sufficient amount of training data from different platforms, this model can possibly present a breakthrough in analysis of discourse in dialogues.
\par We used our model to characterize a small stream of discussions regarding a single topic; that too in a single platform. But this revealed the potential of discourse relation to be exploited in social network analyses. This can be further extended to understand argumentation, particularly in multimodal environment, and how different information sources tend to shape online discussions.
\bibliographystyle{spmpsci}
\bibliography{ref}

\begin{thebibliography}{10}
\providecommand{\url}[1]{{#1}}
\providecommand{\urlprefix}{URL }
\expandafter\ifx\csname urlstyle\endcsname\relax
  \providecommand{\doi}[1]{DOI~\discretionary{}{}{}#1}\else
  \providecommand{\doi}{DOI~\discretionary{}{}{}\begingroup
  \urlstyle{rm}\Url}\fi

\bibitem{arguello2015predicting}
Arguello, J., Shaffer, K.: Predicting speech acts in mooc forum posts.
\newblock In: ICWSM, pp. 2--11 (2015)

\bibitem{bhatia2016identifying}
Bhatia, S., Biyani, P., Mitra, P.: Identifying the role of individual user
  messages in an online discussion and its use in thread retrieval.
\newblock Journal of the Association for Information Science and Technology
  \textbf{67}(2), 276--288 (2016)

\bibitem{bollen2011twitter}
Bollen, J., Mao, H., Zeng, X.: Twitter mood predicts the stock market.
\newblock Journal of computational science \textbf{2}(1), 1--8 (2011)

\bibitem{bunt2010methodology}
Bunt, H.: A methodology for designing semantic annotation languages exploring
  semantic-syntactic isomorphisms.
\newblock In: Proceedings of the Second International Conference on Global
  Interoperability for Language Resources (ICGL 2010), Hong Kong, pp. 29--46
  (2010)

\bibitem{chakraborty2017metrics}
Chakraborty, T., Dalmia, A., Mukherjee, A., Ganguly, N.: Metrics for community
  analysis: A survey.
\newblock ACM Computing Surveys (CSUR) \textbf{50}(4), 54 (2017)

\bibitem{clark2004multi}
Clark, A., Popescu-Belis, A.: Multi-level dialogue act tags.
\newblock In: SIGDIAL, pp. 163--170. Cambridge, MA (2004)

\bibitem{cohen2004learning}
Cohen, W.W., Carvalho, V.R., Mitchell, T.M.: Learning to classify email into
  ``speech acts''.
\newblock In: Proceedings of the 2004 Conference on Empirical Methods in
  Natural Language Processing, pp. 1--8 (2004)

\bibitem{ding2008using}
Ding, S., Cong, G., Lin, C.Y., Zhu, X.: Using conditional random fields to
  extract contexts and answers of questions from online forums.
\newblock Proceedings of ACL-08: HLT pp. 710--718 (2008)

\bibitem{du2017stance}
Du, J., Xu, R., He, Y., Gui, L.: Stance classification with target-specific
  neural attention networks.
\newblock pp. 3988--3994. International Joint Conferences on Artificial
  Intelligence (2017)

\bibitem{dutta2017dialogue}
Dutta, S., Das, D.: Dialogue modelling in multi-party social media
  conversation.
\newblock In: International Conference on Text, Speech, and Dialogue, pp.
  219--227. Springer (2017)

\bibitem{eisenlauer2014facebook}
Eisenlauer, V.: Facebook as a third author—(semi-) automated participation
  framework in social network sites.
\newblock Journal of pragmatics \textbf{72}, 73--85 (2014)

\bibitem{hasan2014you}
Hasan, K.S., Ng, V.: Why are you taking this stance? identifying and
  classifying reasons in ideological debates.
\newblock In: Proceedings of the 2014 Conference on Empirical Methods in
  Natural Language Processing (EMNLP), pp. 751--762 (2014)

\bibitem{hochreiter2001gradient}
Hochreiter, S., Bengio, Y., Frasconi, P., Schmidhuber, J., et~al.: Gradient
  flow in recurrent nets: the difficulty of learning long-term dependencies
  (2001)

\bibitem{hochreiter1997long}
Hochreiter, S., Schmidhuber, J.: Long short-term memory.
\newblock Neural computation \textbf{9}(8), 1735--1780 (1997)

\bibitem{kalchbrenner2013recurrent}
Kalchbrenner, N., Blunsom, P.: Recurrent convolutional neural networks for
  discourse compositionality.
\newblock arXiv preprint arXiv:1306.3584  (2013)

\bibitem{lai2016friends}
Lai, M., Far{\'\i}as, D.I.H., Patti, V., Rosso, P.: Friends and enemies of
  clinton and trump: using context for detecting stance in political tweets.
\newblock In: Mexican International Conference on Artificial Intelligence, pp.
  155--168. Springer (2016)

\bibitem{larson1984meaning}
Larson, M.L.: Meaning-based translation: A guide to cross-language equivalence.
\newblock University press of America Lanham (1984)

\bibitem{le2014distributed}
Le, Q., Mikolov, T.: Distributed representations of sentences and documents.
\newblock In: International Conference on Machine Learning, pp. 1188--1196
  (2014)

\bibitem{lotan2011mapping}
Lotan, G.: Mapping information flows on twitter.
\newblock In: The Future of the Social Web (2011)

\bibitem{malhotra2015long}
Malhotra, P., Vig, L., Shroff, G., Agarwal, P.: Long short term memory networks
  for anomaly detection in time series.
\newblock In: European Symposium on Artificial Neural Networks, Computational
  Intelligence and Machine Learning, pp. 89--94. Presses universitaires de
  Louvain (2015)

\bibitem{mikolov2013distributed}
Mikolov, T., Sutskever, I., Chen, K., Corrado, G.S., Dean, J.: Distributed
  representations of words and phrases and their compositionality.
\newblock In: Advances in neural information processing systems, pp. 3111--3119
  (2013)

\bibitem{misra2017topic}
Misra, A., Walker, M.: Topic independent identification of agreement and
  disagreement in social media dialogue.
\newblock arXiv preprint arXiv:1709.00661  (2017)

\bibitem{o2010tweets}
O'Connor, B., Balasubramanyan, R., Routledge, B.R., Smith, N.A., et~al.: From
  tweets to polls: Linking text sentiment to public opinion time series.
\newblock ICWSM \textbf{11}(122-129), 1--2 (2010)

\bibitem{scott2015pragmatics}
Scott, K.: The pragmatics of hashtags: Inference and conversational style on
  twitter.
\newblock Journal of Pragmatics \textbf{81}, 8--20 (2015)

\bibitem{somasundaran2009supervised}
Somasundaran, S., Namata, G., Wiebe, J., Getoor, L.: Supervised and
  unsupervised methods in employing discourse relations for improving opinion
  polarity classification.
\newblock In: Proceedings of the 2009 Conference on Empirical Methods in
  Natural Language Processing: Volume 1-Volume 1, pp. 170--179. Association for
  Computational Linguistics (2009)

\bibitem{suh2010want}
Suh, B., Hong, L., Pirolli, P., Chi, E.H.: Want to be retweeted? large scale
  analytics on factors impacting retweet in twitter network.
\newblock In: 2010 IEEE second international conference on Social computing
  (socialcom), pp. 177--184. IEEE (2010)

\bibitem{sundermeyer2012lstm}
Sundermeyer, M., Schl{\"u}ter, R., Ney, H.: Lstm neural networks for language
  modeling.
\newblock In: Thirteenth Annual Conference of the International Speech
  Communication Association (2012)

\bibitem{sutskever2014sequence}
Sutskever, I., Vinyals, O., Le, Q.V.: Sequence to sequence learning with neural
  networks.
\newblock In: Advances in neural information processing systems, pp. 3104--3112
  (2014)

\bibitem{tan2011user}
Tan, C., Lee, L., Tang, J., Jiang, L., Zhou, M., Li, P.: User-level sentiment
  analysis incorporating social networks.
\newblock In: Proceedings of the 17th ACM SIGKDD international conference on
  Knowledge discovery and data mining, pp. 1397--1405. ACM (2011)

\bibitem{trevithick2008method}
Trevithick, P., Clippinger, J.H.: Method and system for characterizing
  relationships in social networks (2008).
\newblock US Patent 7,366,759

\bibitem{wang2012system}
Wang, H., Can, D., Kazemzadeh, A., Bar, F., Narayanan, S.: A system for
  real-time twitter sentiment analysis of 2012 us presidential election cycle.
\newblock In: Proceedings of the ACL 2012 System Demonstrations, pp. 115--120.
  Association for Computational Linguistics (2012)

\bibitem{wang2015bandwagon}
Wang, K.C., Lai, C.M., Wang, T., Wu, S.F.: Bandwagon effect in facebook
  discussion groups.
\newblock In: Proceedings of the ASE BigData \& SocialInformatics 2015, p.~17.
  ACM (2015)

\bibitem{wang2016attention}
Wang, Y., Huang, M., Zhao, L., et~al.: Attention-based lstm for aspect-level
  sentiment classification.
\newblock In: Proceedings of the 2016 Conference on Empirical Methods in
  Natural Language Processing, pp. 606--615 (2016)

\bibitem{wong2013media}
Wong, F., Tan, C.W., Sen, S., Chiang, M.: Media, pundits and the us
  presidential election: Quantifying political leanings from tweets.
\newblock In: Proceedings of the International Conference on Weblogs and Social
  Media, pp. 640--649 (2013)

\bibitem{zhang2017characterizing}
Zhang, A.X., Culbertson, B., Paritosh, P.: Characterizing online discussion
  using coarse discourse sequences.
\newblock In: Proceedings of the Eleventh International Conference on Web and
  Social Media. AAAI Press, pp. 1--10 (2017)

\end{thebibliography}
\end{document}